\documentstyle[latex-acl,psfig]{article}

\title{\vspace{-0.5in}\LARGE\bf COMPACT REPRESENTATIONS BY
FINITE-STATE TRANSDUCERS}
\author{Mehryar Mohri\\
Institut Gaspard Monge-LADL\\
Universit\'e Marne-la-Vall\'ee\\
2, rue de la Butte verte\\
93160 Noisy-le-Grand, FRANCE
\\Internet: mohri@univ-mlv.fr\\}

\begin{document}

\maketitle
\vspace{-0.5in}
\begin{abstract}

Finite-state transducers give efficient representations of many
Natural Language phenomena. They allow to account for complex lexicon
restrictions encountered, without involving the use of a large set of
complex rules difficult to analyze. We here show that these
representations can be made very compact, indicate how to perform the
corresponding minimization, and point out interesting linguistic
side-effects of this operation.

\end{abstract}

\section*{1. MOTIVATION}

Finite-state transducers constitute appropriate representations of
Natural Language  phenomena. Indeed, they have been shown to be
sufficient tools to describe morphological and phonetic forms of a
language (Karttunen et al., 1992; Kay and Kaplan, 1994). Transducers
can then be viewed as functions which map lexical representations to
the surface forms, or inflected forms to their phonetic
pronunciations, and vice versa. They allow to avoid the use of a
great set of complex rules often difficult to check, handle, or even
understand.

Finite-state automata and transducers can also be used to represent
the syntactic constraints of languages such as English or French
(Koskenniemi, 1990; Mohri, 1993; Pereira, 1991; Roche, 1993). The
syntactic analysis can then be reduced to performing the intersection
of two automata, or to the application of a transducer to an
automaton. However, whereas first results show that the size of the
syntactic transducer exceeds several hundreds of thousands of states,
no upper bound has been proposed for it, as the representation of all
syntactic entries has not been done yet. Thus, one may ask whether
such representations could succeed on a large scale.

It is therefore crucial to control or to limit the size of these
transducers in order to avoid a blow up. Classic minimization
algorithms permit to reduce to the minimal the size of a
deterministic automaton recognizing a given language (Aho et al.,
1974). No similar algorithm has been proposed in the case of
sequential transducers, namely transducers whose associated input
automata are deterministic.

We here briefly describe an algorithm which allows to compute a
minimal transducer, namely one with the least number of states, from
a given subsequential transducer. In addition to the desired property
of minimization, the transducer obtained in such a way has
interesting linguistic properties that we shall indicate. We have
fully implemented and experimented this algorithm in the case of
large scale dictionaries. In the last section, we shall describe
experiments and corresponding results. They show this algorithm to be
very efficient.

\section*{2. ALGORITHM}

Our algorithm can be applied to any sequential transducer $T= (V, i,
F, A, B, \delta, \sigma)$ where: $V$ is the set of the states of $T$,
$i$ its initial state, $F$ the set of its final states, $A$ and $B$
respectively the input and output alphabet of the transducer,
$\delta$ the state transition function which maps $V \times A$ to
$V$, and $\sigma$ the output function which maps $V \times A$ to
$B^*$. With this definition, input labels are elements of the
alphabet, whereas output labels can be words. Figure 1 gives an
example of a sequential transducer.

Transducers can be considered as automata over the alphabet $A \times
B^*$. Thus, considered as such they can be submitted to the
minimization in the sense of automata. Notice however that the
application of the minimization algorithm for automata does not
permit to reduce the number of states of the transducer $T$. We shall
describe in the following how the algorithm we propose allows to
reduce the number of states of this transducer.

\begin{figure}
\begin{picture}(250,100)(0,100)
\put(250,130){\makebox(0,0){\psfig{figure=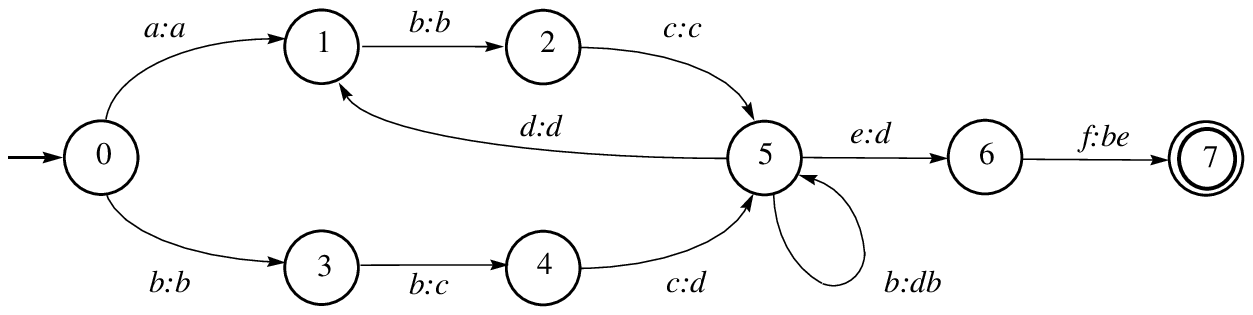}}}
\makebox(480,100){Figure 1. Transducer $T$.}
\end{picture}
\vspace{1,5 cm}
\end{figure}

This algorithm works in two stages. The first one modifies only the
output automaton associated with the given sequential transducer $T$.
Thus, we can denote by $(V, i, F, A, B, \delta, \sigma_2)$ the
transducer $T_2$ obtained after this first stage. Let $P$ be the
function which maps $V$ to $B^*$ which associates with each state $q$
of $T$ the greatest common prefix of all the words which can be read
on the output labels of $T$ from $q$ to a final state. The value of
$P(5)$ is for instance $db$ since this is the greatest common prefix
of the labels of all output paths leaving $3$. In particular, if $q$
is a final state then $P(q)$ is the empty word $\epsilon$. In order
to simplify this presentation, we shall assume in the following that
$P(i)=\epsilon$. The output function $\sigma_2$ of $T_2$ is defined
by:\\ \\
\begin{tabular}{ll}
$\forall q \in V,$ & $\forall a \in A,$ \\
&$ \sigma_2(q, a) = (P(q))^{-1}\sigma(q, a)P(\delta(q,a)).$
\end{tabular} \\ \\
Namely, the output labels of $T$ are modified in such a way that they
include every letter which would necessarily be read later on the
following transitions. Figure 2 illustrates these modifications.

\begin{figure}[hb]
\begin{picture}(250,100)(0,100)
\put(250,140){\makebox(0,0){\psfig{figure=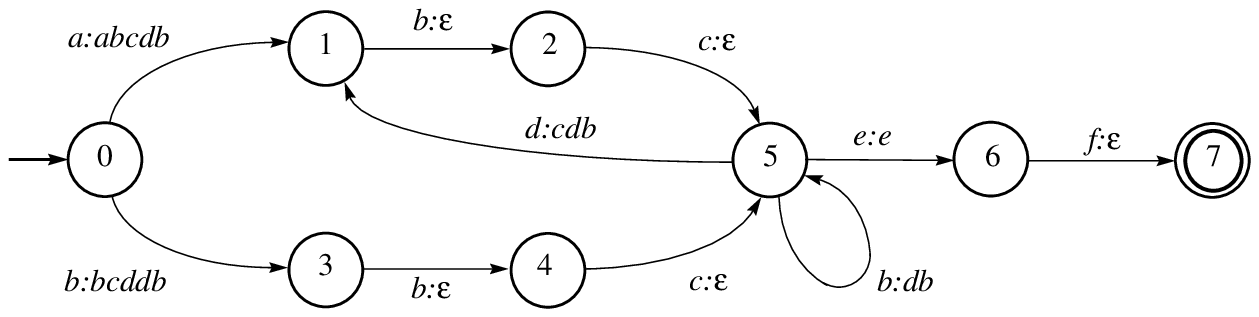}}}
\makebox(480,110){Figure 2. Transducer $T_2$.}
\end{picture}
\vspace{1,5 cm}
\end{figure}

It shows the transducer $T_2$ obtained from $T$ by performing the
operations described above. Notice that only the output labels of $T$
have been modified. The output label $a$ corresponding to the
transition linking states $0$ and $1$ of the transducer has now
become $abcdb$ as this is the longest word which is necessarily read
from the initial state $0$ of

\begin{figure}
\begin{picture}(250,100)(0,100)
\makebox(480,100){}
\end{picture}
\vspace{1,5 cm}
\end{figure}
$T$ if beginning with the transition $(0,1)$. The output label of the
following transition of $T_2$ is now empty. Indeed, anything which
could be read from the transition $(1,2)$ on the output labels has
now been included in the previous transition $(0,1)$.

It is easy to show that the transducer $T_2$ obtained after the first
stage is equivalent to $T$. Namely, these two transducers correspond
to the same function mapping $A^*$ to $B^*$. One may notice, however,
that unlike $T$ this transducer can be minimized in the sense of
automata and that this leads to a transducer with only six states.
Figure 3 indicates the transducer $T_3$ obtained in such a way.

The second stage of our algorithm precisely consists of the
application of the minimization in the sense of automata, that is, of
merging equivalent states of the transducer. It can be showed that
the application of the two presented stages \begin{figure}[hb]
\begin{picture}(250,100)(0,100)
\put(250,140){}
\end{picture}
\vspace{1,5 cm}
\end{figure}to a sequential transducer $T$ systematically leads to an
equivalent sequential transducer with the minimal number of states
(Mohri, 1994). Indeed, the states of this minimal transducer can be
characterized by the following equivalence relation: two states of a
sequential transducer are equivalent if and only if one can read the
same words from these states using the left automaton associated with
this transducer (equivalence in the sense of automata) and if the
corresponding outputs from these states differ by the same prefix for
any word leading to a final state. Thus, the described algorithm can
be considered as optimal.

Notice that we here only considered sequential transducers, but not
all transducers representing sequential functions are sequential.
However, transducers which are not sequential though representing a
sequential function can be determinized using a procedure close to
the one used for the determinization of automata. The algorithm above
can then be applied to such determinized transducers.

The complexity of the application of a non sequential transducer to a
string is not linear. This is not the case even for non-deterministic
automata. Indeed, recognizing a word $w$ with a non-deterministic
automaton of $|V|$ states each containing at most $e$ leaving
transitions requires $O(e|V||w|)$ (see Aho et al., 1974). The
application of a non-sequential transducer is even more time
consuming, so the determinization of transducers clearly improves
their application. We have considered above sequential transducers,
but transducers can be used in two ways. These transducers, although
they allow linear time application on left, are generally not
sequential considered as right input transducers. However, the first
stage of the presented algorithm constitutes a pseudo-determinization
of right input transducers. Indeed, as right labels (outputs) are
brought closer to the initial state as much as possible, irrelevant
paths are sooner rejected.

\begin{figure}
\begin{picture}(250,100)(0,100)
\put(250,140){\makebox(0,0){\psfig{figure=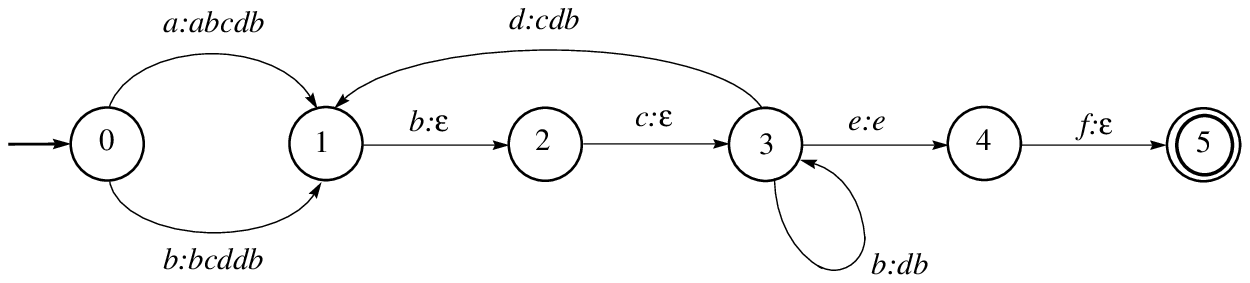}}}
\makebox(480,140){Figure 3. Transducer $T_3$.}
\end{picture}
\vspace{0,5 cm}
\end{figure}

Consider for example the string $x=abcdbcdbe$ and compare the
application of transducers $T$ and $T_2$ to this sequence on right
input. Using the transducer $T$, the first three letters of this
sequence lead to the single state 5, but then reading $db$ leads to a
set of states $\{1,5,6\}$. Thus, in order to proceed with the
recognition, one needs to store this set and consider all possible
transitions or paths from its states. Using the transducer $T_2$ and
reading $abcdb$ give the single state ${1}$. Hence, although the
right input transducer is not sequential, it still permits to reduce
the num\begin{figure}
\begin{picture}(250,100)(0,100)
\put(250,140){}
\makebox(480,140){}
\end{picture}
\vspace{0,5 cm}
\end{figure}
ber of paths and states to visit. This can be considered as another
advantage of the method proposed for the minimization of sequential
transducers: not only the transducer is sequential and minimal on one
side, but it is also pseudo-sequential on the other side.

The representation of language often reveals ambiguities. The
sequential transducers we have just described do not allow them.
However, real ambiguities encountered in Natural Language Processing
can be assumed to be finite and bounded by an integer $p$. The use of
the algorithm above can be easily extended to the case of
subsequential transducers and even to a larger category of
transducers which can represent ambiguities and which we shall call
{\em $p$-subsequential transducers}. These transducers are provided
with $p$ final functions $\phi_i$, ($ i \in [1,p]$) mapping $F$, the
set of final states, to $B^*$. Figure 4 gives an example of a
$2$-subsequential transducer.

\begin{figure}[hb]
\begin{picture}(250,30)(0,30)
\put(125,40){\makebox(0,0){\psfig{figure=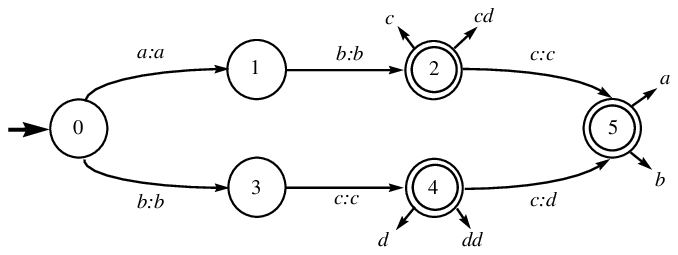}}}
\makebox(230,-60){Figure 4. $2$-subsequential transducer $T_4$.}
\end{picture}
\vspace{2 cm}

\end{figure}

The application of these transducers to a string $x$ is similar to
the one generally used for sequential ones. It outputs a string
corresponding to the concatenation of consecutive labels encoutered.
However, the output string obtained once reaching state $q$ must here
be completed by the $\phi _i (q)$ without reading any additional
input letter. The application of the transducer $T_{4}$ to the word
$abc$ for instance provides the two outputs $abca$ and $abcb$.

The extension of the use of the algorithm above is easy. Indeed, in
all cases $p$-subsequential transducers can be transformed into
sequential transducers by adding $p$ new letters to the alphabet $A$,
and by replacing the $p$ final functions by transitions labeled with
these new letters on input and the corresponding values of the
functions on output. These transitions would leave the final states
and reach a newly created state which would become the single final
state of the transducer. The minimal transducer associated with the
$2$-subsequential transducer $T_4$ is shown on figure 5. It results
from $T_{4}$ by merging the states $2$ and $4$ after the first stage
of pseudo-determinization.
\vspace{0,75 cm}
\begin{figure}[hb]
\begin{picture}(250,40)(0,0)
\put(125,40){\makebox(0,0){\psfig{figure=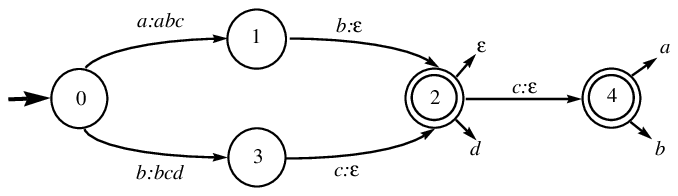}}}
\makebox(220,-60){Figure 5. Minimal $2$-subsequential transducer
$T_5$.}
\end{picture}
\vspace{1 cm}
\end{figure}

In the following section, we shall describe some of the experiments
we carried out and the corresponding results. These experiments use
the notion of $p$-subsequential transducers just developped as they
all deal with cases where ambiguities appear.

\section*{3. EXPERIMENTS, RESULTS, AND PROPERTIES}
\label{EXPERIMENTS, RESULTS, AND PROPERTIES}
We have experimented the algorithm described above by applying it to
several large scale dictionaries. We have applied it to the
transducer which associates with each French word the set of its
phonetic pronunciations. This transducer can be built from a
dictionary (DELAPF) of inflected forms of French, each followed by
its pronunciations (Laporte, 1988). It can be easily transformed into
a sequential or $p$-subsequential transducer, where $p$, the maximum
number of ambiguities for this transducer, is about four (about 30
words admit 4 different pronunciations). This requires that the
transducer be kept deterministic while new associations are added to
it.

The dictionary contains about 480.000 entries of words and phonetic
pronunciations and its size is about 10 Mb. The whole minimization
algorithm, including building the transducer from the dictionary and
the compression of the final transducer, was quite fast: it took
about 9 minutes using a HP 9000/755 with 128 Mb of RAM. The resulting
transducer contains about 47.000 states and 130.000 transitions.
Since it is sequential, it can be better compressed as one only needs
to store the set of its transitions. The minimal transducer obtained
has been put in a compact form occupying about 1,1 Mb. Also, as the
transducer is sequential, it allows faster recognition times.

In addition to the above results, the transducer obtained by this
algorithm has interesting properties. Indeed, when applied to an
input word $w$ which may not be a French word this transducer outputs
the longest common prefix of the phonetic transcriptions of all words
beginning with $w$. The input $w=opio$ for instance, though it does
not constitute a French word, yields {\em opjoman}. Also, $w=opht$
gives {\em oftalm}. This property of minimal transducers as defined
above could be used in applications such as OCR or spellchecking, in
order to restore the correct form of a word from its beginning, or
from the beginning of its pronunciation. \\

              Table 1. Results of minimization experiments \\
\begin{tabular}{||l|c|c|c||} \hline
& DELAPF&FDELAF& EDELAF \\ \hline
Initial size & 9,6 Mb & 22,3 Mb & 3,5 Mb \\ \hline
Entries & 480.000 & 780.000 & 145.000 \\ \hline
Max. ambg & 4 & 15 & 8 \\ \hline
Final size& 1,1 Mb & 1,6 Mb & 1 Mb \\ \hline
States & 47.000 & 66.000 & 47.000 \\ \hline
Transitions & 130.000 & 195.000 & 115.000 \\ \hline
Alphabet& 13.500 & 20.000 & 14.000 \\ \hline
Time spent & 9' & 20' & 7' \\ \hline
\end{tabular} \\ \\

We have also performed the same experiment using 2 other large
dictionaries: French (FDELAF) (Courtois, 1989) and English (EDELAF)
(Klarsfeld, 1991) dictionaries of inflected forms. These dictionaries
are made of associations of inflected forms and their corresponding
canonical representations. It took about 20 minutes constructing the
$15$-subsequential transducer associated with the French dictionary
of about 22 Mb. Here again, properties of the obtained transducers
seem interesting for various applications. Given the input {\em
w=transducte} for instance the transducer provides the output {\em
transducteur.N1:m}. Thus, although $w$ is not a correct French word,
it provides two additional letters completing this word, and
indicates that it is a masculine noun. Notice that no information is
given about the number of this noun as it can be completed by an
ending $s$ or not. Analogous results were obtained using the English
dictionary. A part of them is illustrated by the table above. It
allows to compare the initial size of the file representing these
dictionaries and the size of the equivalent transducers in memory
(final size). The third line of the table gives the maximum number of
lexical ambiguities encountered in each dictionary. The following
lines indicate the number of states and transitions of the
transducers and also the size of the alphabet needed to represent the
output labels. These experiments show that this size remains small
compared to the number of transitions. Hence, the use of an
additional alphabet does not increase noticeably the size of the
transducer. Also notice that the time indicated corresponds to the
entire process of transformation of the file dictionaries into
tranducers. This includes of course the time spent for I/O's. We have
not tried to optimize these results. Several available methods should
help both to reduce the size of the obtained transducers and the time
spent for the algorithm.

\section*{4. CONCLUSION}
\label{CONCLUSION}

We have informally described an algorithm which allows to compact
sequential transducers used in the description of language.
Experiments on large scale dictionaries have proved this algorithm to
be efficient. In addition to its use in several applications, it
could help to limit the growth of the size of the representations of
syntactic constraints.

\section*{REFERENCES}

Aho, Alfred, John Hopcroft, Jeffery Ullman. 1974. {\em The design and
analysis of computer algorithms}. Reading, Mass.: Addison Wesley.\\

Courtois, Blandine. 1989. {\em DELAS: Dictionnaire Electronique du
LADL pour les mots simples du fran[ais} Technical Report, LADL,
Paris, France. \\

Karttunen, Laura, Ronald M.~Kaplan, and Annie Zaenen. 1992. Two-level
Morphology with Composition. {\em Proceedings of the fifteenth
International Conference on Computational Linguistics {(COLING'92)}},
Nantes, France, August.\\

Kay, Martin, and Ronald M.~Kaplan. 1994. Regular Models of
Phonological Rule Systems. To appear in {\em Computational
Linguistics}.\\

Klarsfeld, Gaby. 1991. {\em Dictionnaire morphologique de l'anglais}.
Technical Report, LADL, Paris, France. \\

Koskenniemi Kimmo. 1990. Finite-state Parsing and Disambiguation.
{\em Proceedings of the thirteenth International Conference on
Computational Linguistics {(COLING'90)}}, Helsinki, Finland.\\

Laporte, Eric. 1988. {\em M\'ethodes algorithmiques et lexicales de
phon\'etisation de textes.} Ph.D thesis, Universit\'e Paris 7, Paris,
France.\\

Mohri, Mehryar. 1993. {\em Analyse et repr\'esentation par automates
de structures syntaxiques compos\'ees.} Ph.D thesis, Universit\'e
Paris 7, Paris, France.\\

Mohri, Mehryar. 1994. Minimization of Sequential Transducers. {\em
Proceedings of Combinatorial Pattern Matchnig (CPM'94)},
Springer-Verlag, Berlin Heidelberg New York. Also Submitted to {\em
Theoretical Computer Science}.\\

Pereira, Fernando C.~N. 1991. Finite-State Approximation of Phrase
Structure Grammars. {\em Proceedings of the 29th Annual Meeting of
the Association for Computational Linguistics {(ACL'91)}}, Berkeley,
California.\\

Roche Emmanuel. 1993. {\em Analyse syntaxique transformationnelle du
fran\c{c}ais par transducteur et lexique-grammaire.} Ph.D thesis,
Universit\'e Paris 7, Paris, France.\\

\end{document}